\documentclass[12pt]{article}
\usepackage[latin9]{inputenc}
\usepackage{units}
\usepackage{amsmath}
\usepackage{amssymb}
\usepackage{amsthm}
\usepackage{overpic}
\usepackage{multirow}
\usepackage{rotating}
\usepackage{paralist}
\usepackage[usenames,dvipsnames]{color}

\makeatletter

\usepackage{graphicx} 

\usepackage{amsfonts}
\usepackage{algorithmic}\usepackage{algorithm}
\newtheorem{theo}{Theorem}[section]

\newcommand{\bb}[1]{\boldsymbol{\mathrm{#1}}}

\def\Tr{\top}

\newcommand{\RR}{\mathbb{R}}

\newcommand{\uu}{\bb{u}}

\newcommand{\Ff}{\bb{F}}

\newcommand{\Aa}{\bb{A}}
\newcommand{\Bb}{\bb{B}}

\newcommand{\Uu}{\bb{U}}
\newcommand{\Dd}{\bb{D}}

\newcommand{\Ww}{\bb{W}}
\newcommand{\Ll}{\bb{L}}

\newcommand{\tr}{\mathrm{tr}\,}

\makeatother

\begin{document}

\title{Heat kernel coupling for multiple graph analysis}

\author{Michael M. Bronstein \and Klaus Glashoff\\
\small Institute of Computational Science, Faculty of Informatics,\\
\small Universit{\`a} della Svizzera Italiana, Lugano, Switzerland 
}
\maketitle
\begin{abstract}
In this paper, we introduce heat kernel coupling (HKC)  as a method of constructing multimodal spectral geometry on weighted graphs of different size without vertex-wise bijective correspondence. We show that Laplacian averaging can be derived as a limit case of HKC, and demonstrate its applications on several problems from the manifold learning and pattern recognition domain.  
\end{abstract}

\section{Introduction}

%
%
%
%
%
%
Many problems in computer vision and pattern recognition boil down to constructing a Laplacian operator describing some data manifold and  finding its eigenvectors. Notable examples include spectral clustering~\cite{Ng01onspectral},  
eigenmaps~\cite{Belkin02laplacianeigenmaps}, diffusion maps and distances \cite{Coifman05geometricdiffusions,Nadler05diffusionmaps}, 
spectral graph partitioning~\cite{DingHeetal2001}, 
spectral hashing~\cite{weiss2008spectral},   
and image segmentation~\cite{Shi97normalizedcuts}. 

%

Recently, there has been an increased interest in extending spectral geometric constructions to the multimodal setting, involving two or more data spaces. 
Many data analysis applications involve observations and measurements of data 
using different modalities, such as multimedia documents~\cite{bekkerman2005multi,weston2010large,rasiwasia2010new,mcfee2011learning,sfchang13,masci}, audio and video~\cite{kidron2005pixels,alameda2011finding,sharma2012generalized}, 
images with different lighting conditions \cite{bansaljoint}, 
or medical imaging modalities \cite{bronstein2010data}.  

Multimodal (or `multi-view') clustering was studied in the computer vision and pattern recognition community \cite{de2005spectral,MaAnchor2008,tang2009clustering,Cai2011,kumarco,dong2013clustering}.  
%
Sindhwani et al. \cite{sindhwani2005co} used a convex combination of Laplacians in the `co-regularization' framework.  
{\em Manifold alignment} considered multiple manifolds as a single space with `connections' between points and tries to find an aligned set of eigenvectors \cite{ham2005semisupervised,wangageneral2009,wang2008manifold}. 
Eynard et al. \cite{eynard2012multimodal} proposed finding a common eigenbasis of multiple Laplacians by means of joint approximate diagonalization (JADE). 
%
Kovnatsky et al. \cite{KovBBKK:2013:EG} improved this method using subspace parametrization. 
Bronstein et al. \cite{cco} studied the problem of finding closest commuting operators (CCO) and showed its equivalence to joint diagonalization.

One of the main limitations of JADE and CCO problems is the assumption of given bijective correspondence (or more generally, functional correspondence \cite{ovsjanikov2012functional}) between the underlying manifolds or graphs. 
In this paper, we consider the setting where such correspondence is unknown or may not exist, and instead, one is given a set of corresponding functions. We show a problem similar to CCO, wherein we try to minimally modify the Laplacians such that the corresponding heat kernels behave consistently. 
In the limit case with given bijective correspondence, this heat kernel coupling problem is equivalent to Laplacian averaging.

\section{Background}

{\bf Notation and definitions. }
Let $\Aa, \Bb$ be two $n\times n$ real symmetric matrices. We denote by 
\begin{eqnarray*}
\| \Aa \|_\mathrm{F} &=& \textstyle \left( \sum_{ij}|a_{ij}|^2\right)^{1/2} = \left( \mathrm{tr}(\Aa^\Tr \Aa) \right)^{1/2} 
\end{eqnarray*} 
the {\em Frobenius} 
norm 
of $\Aa$. 
We say that $\Aa$ and $\Bb$ {\em commute} if $\Aa\Bb = \Bb\Aa$, and call $[\Aa,\Bb] = \Aa\Bb - \Bb\Aa$ their {\em commutator}. 
If there exists a unitary matrix $\hat{\bb{U}}$ such that $\hat{\bb{U}}^\Tr \Aa \hat{\bb{U}} = \bb{\Lambda}_A$ and $\hat{\bb{U}}^\Tr \Bb \hat{\bb{U}} = \bb{\Lambda}_B$ are diagonal, we say that $\Aa, \Bb$ are {\em jointly diagonalizable} and call such $\hat{\bb{U}}$ the {\em joint eigenbasis} of $\Aa$ and $\Bb$. 
Two matrices are jointly diagonalizable iff they commute.

We denote by $\mathrm{diag}(\Aa)$ a column vector containing the diagonal elements of matrix $\Aa$, and by $\mathrm{diag}(a_1,\hdots, a_n)$ a diagonal matrix containing on the diagonal the elements $a_1, \hdots, a_n$. Furthermore, we use $\mathrm{Diag}(\Aa) = \mathrm{diag}(\mathrm{diag}(\Aa))$ to denote a diagonal matrix obtained by setting to zero the off-diagonal elements of $\Aa$.

{\bf Laplacians.}
Let us be given an undirected weighted graph $G = (V,E,\bb{W})$ without loops (i.e., a \emph{simple graph} ) with vertice set $V = \{x_1, \hdots, x_n\}$ and edges $E \subseteq ,\{1,\hdots, n \}^2$ such that $(i,i)\notin E$ for $i=1,\hdots,n$. Let $E=\{(i_k,j_k) | k=1,\hdots ,|E|\}$. There are given non-negative weights $w_{ij} \geq 0$, satisfying $w_{ij} = 0$ if $x_i, x_j$ are not connected (i.e., $(i,j) \notin E$).  
The $n\times n$ matrix $\Ww = (w_{ij})$ is called the {\em adjacency matrix}  and 
\begin{eqnarray}
\Ll = \Dd - \Ww, \hspace{5mm} \Dd = \mathrm{diag}\left(\sum_{j\neq 1} w_{1j}, \hdots, \sum_{j\neq n} w_{nj} \right)
\label{eq:laplacian}
\end{eqnarray} 

Hereinafter, we denote by $\mathcal{L}(V,E)$  
the set of all \emph{valid Laplacian} matrices of a simple graph $(V,E)$, which is defined as follows: $L=(l_{ij})\in \mathcal{L}(V,E)$ \emph{iff}

\begin{inparaenum}[(i)] 
\item 
$l_{ij} \leq 0$  and $l_{ij} = l_{ji}$ for $i\neq j$;
\item sparse structure: $l_{ij} = 0$ if $(i,j) \notin E$; and 
\item zero row sum: $\sum_{j=1}^n l_{ij} = 0$. 
\end{inparaenum}
Defining the Laplacian according to~(\ref{eq:laplacian}) through the edge weight matrix $\Ww$, we automatically get properties (i) - (iii) satisfied. The other way round: Any valid Laplacian of a simple graph - in the sense of (i)-(iii) - gives rise to a weight matrix  $\Ww$ of a simple weighted graph by defining $w_{ij}=-l_{ij}, i\neq j$.

For numerical purposes, we will make use of a proper \emph{parametrization} of the set of valid Laplacians. Let $m=|E|$ denote the number of edges of $G$. For $\uu = (u_1,\hdots,u_m)\in \mathbb{R}_{\geq 0}^m$ we define the weight matrix $\Ww$ by
 
\begin{eqnarray}
w_{ij}(\uu) &=& \left\{
\begin{array}{cc}
u_l & i = i_l, j = j_l \,\, \text{or} \,\, i = j_l, j = i_l \\
0 & \text{else}.
\end{array}
\right. 
\label{eq:wparam}
\end{eqnarray}

Defining $\Ll$ as in (\ref{eq:laplacian}), the requirements (i)-(iii)of a valid Laplacian are satisfied.
In undirected weighted graph, the matrices $\bb{W}$ and $\bb{L}$ are symmetric. Furthermore, $\Ll$ is positive semi-definite. 
Consequently, $\Ll $ admits the unitary eigendecomposition $\bb{L} = \bb{\Phi} \bb{\Lambda} \bb{\Phi}^\Tr$ with orthonormal eigenvectors $\bb{\Phi} = (\bb{\phi}_1,\hdots, \bb{\phi}_n)$ and real eigenvalues $0=\lambda_1 \leq \lambda_2 \leq \hdots \leq \lambda_n$, $\bb{\Lambda} = \mathrm{diag}(\lambda_1,\hdots, \lambda_n)$.

{\bf Heat diffusion on graphs.} 
Let $f: V \rightarrow \RR$ denote a function defined on the vertex set of the graph. We can identify $f$ with an $|V|$-dimensional vector $\bb{f} = (f(x_1),\hdots, f(x_{|V|}))$, and denote by $\mathcal{F}(V)$ the space of such functions. 
%

Similarly to the standard heat diffusion equation, one can define a diffusion process on $G$, governed by the following equation: 
\begin{equation}
\bb{L} \bb{f}(t) + \frac{\partial}{\partial t}\bb{f}(t) = 0, \hspace{5mm} \bb{f}(0) = \bb{f}_0, 
\label{eq:heat}
\end{equation}
where the solution $\bb{f}(t): V \times [0,\infty) \rightarrow \RR_+$ 
is the amount of heat at time $t$ at the vertices $V$.  
The solution of the heat equation is given by $\bb{f}(t) = \bb{H}^t \bb{f}_0$, 
where 
$$\bb{H}^t = e^{-t \Ll} = \bb{\Phi} e^{-t \bb{\Lambda}} \bb{\Phi}^\Tr$$  
is the {\em heat operator} (or the {\em heat kernel}). 

The heat kernel gives rise to the {\em diffusion distance} \cite{Coifman05geometricdiffusions, Coifman} 
\begin{eqnarray}
d_t(x_p,x_q) = \left( \sum_{i=1}^n ((\bb{H}^t)_{pi} - (\bb{H}^t)_{qi})^2  \right)^{1/2} = \left( \sum_{i=1}^n e^{-2t \lambda_i } (\phi_{pi} - \phi_{qi})^2 \right)^{1/2},     
\label{eq:diffdist}
\end{eqnarray}
measuring the `reachability' of vertex $x_q$ from vertex $x_p$ in time $t$.

\section{Multimodal spectral geometry}

Consider two graphs $G_k = (V,E,\bb{W}_k),\, k= 1,2$ with the same vertices $V$ and edges $E$ with different weights $\bb{W}_k$. 
We denote their respective Laplacians by $\bb{L}_k \in \mathcal{L}(V,E)$. 
Such graphs are referred to as {\em multi-level graphs} \cite{dong2013clustering}, and are used to represent multiple modalities or `views' of the same data. \footnote{For simplicity, we consider only two modalities, though extension to more modalities is straightforward.  } 
The main topic of this paper is how to re-define the above spectral geometric constructions (heat kernels, diffusion distances, etc.) in a way that they account for information from both graphs.

\subsection{Laplacian averaging}
The simplest approach 
is to define an average weight $\bar{\bb{W}} = \frac{1}{2}(\bb{W}_1+\bb{W}_2)$ \cite{MaAnchor2008}. 
Equivalently, this problem can be posed as finding a new Laplacian $\bar{\Ll} \in \mathcal{L}(V,E)$ that is equidistant from the given $\Ll_1, \Ll_2$, 
\begin{eqnarray}
\min_{\bar{\Ll} \in L(V,E)} \sum_{i=1}^2 \|\Ll_i - \bar{\Ll} \|_{\mathrm{F}}^2. 
\label{eq:avg}
\end{eqnarray} 
The solution of~(\ref{eq:avg}) is the {\em average Laplacian} $\bar{\Ll} = \frac{1}{2}(\Ll_1 + \Ll_2)$.

\subsection{Joint diagonalization}
Instead of averaging the Laplacians, Eynard et al. \cite{eynard2012multimodal} proposed `averaging' their eigenspaces by means of a joint diagonalization approach: 
construct a common (approximate) eigenbasis $\hat{\bb{\Phi}}$ that (approximately) diagonalizes the Laplacians $\bb{L}_1, \bb{L}_2$, by the following minimization 
\begin{eqnarray}
J(\Ll_1, \Ll_2) = \min_{ \hat{\bb{\Phi}} \in \RR^{n\times n}  } \sum_{k=1}^2\mathrm{off}(\hat{\bb{\Phi}}^\Tr \Ll_k \hat{\bb{\Phi}}) \,\,\,\,\, \text{s.t.} \,\,\,\,\, \hat{\bb{\Phi}}^\Tr \hat{\bb{\Phi}} = \bb{I}, 
\label{eq:jade}
\end{eqnarray}
where $\mathrm{off}(\Aa) = \sum_{i\neq j} a_{ij}^2$ denotes the squared norm of the off-diagonal elements of a matrix \cite{Cardoso96jacobiangles}. 
The joint basis $\hat{\bb{\Phi}}$ obtained in this way 
satisfies 
$\hat{\bb{\Phi}}^\Tr \bb{L}_k \hat{\bb{\Phi}} \approx \mathrm{diag}(\hat{\lambda}_{k,1}, \hdots, \hat{\lambda}_{k,|V|})$. 
The approximate matrices 
$$\hat{\Ll}_k = \hat{\bb{\Phi}}\mathrm{Diag}(\hat{\bb{\Phi}}^\Tr \Ll_k \hat{\bb{\Phi}}) \hat{\bb{\Phi}}^\Tr \approx \Ll_k, $$ 
obtained by setting to zero the off-diagonal elements of $\hat{\bb{\Phi}}^\Tr \bb{L}_k \hat{\bb{\Phi}}$ 
are jointly diagonalizable by $\hat{\bb{\Phi}}$.   
Importantly, in most cases $\hat{\Ll}_k \notin \mathcal{L}(V,E)$, i.e., Laplacian structure does not survive joint diagonalization. 

%


\subsection{Closest commuting operators}
In \cite{cco}, we considered a different problem of finding a pair $\tilde{\Ll}_1, \tilde{\Ll}_2$ of commuting matrices 
(referred to as {\em closest commuting operators} or CCOs) that are closest to the given $\Ll_1, \Ll_2$, 
%
\begin{eqnarray}
C(\Ll_1,\Ll_2) &=& \displaystyle \min_{ \tilde{\Ll}_1, \tilde{\Ll}_1 \in \RR^{n \times n} }  
\sum_{i=1}^2 \| \tilde{\Ll}_i - \Ll_i \|_\mathrm{F}^2  
\,\,\, \text{s.t.} \,\,\, \tilde{\Ll}_1 \tilde{\Ll}_2 = \tilde{\Ll}_2 \tilde{\Ll}_1;  
\label{eq:cco}
\end{eqnarray} 
%
%
and showed that  this problem 
is equivalent to JADE~(\ref{eq:jade}) in the following sense: 

\begin{theo}
Let $\Aa, \Bb$ be symmetric matrices. Then: 

\noindent 1. $C(\Aa,\Bb) = J(\Aa,\Bb)$.  

\noindent 2. Let $\hat{\bb{U}}$ be the approximate joint eigenbasis of $\Aa, \Bb$ obtained by solving the JADE problem~(\ref{eq:jade}). 
Then, 
$\tilde{\Aa} = \hat{\bb{U}}\mathrm{Diag}(\hat{\bb{U}}^\Tr \Aa \hat{\bb{U}}) \hat{\bb{U}}^\Tr$ and  
$\tilde{\Bb} = \hat{\bb{U}}\mathrm{Diag}(\hat{\bb{U}}^\Tr \Bb \hat{\bb{U}}) \hat{\bb{U}}^\Tr$ 
are the closest commuting matrices to $\Aa,\Bb$ solving the CCO problem~(\ref{eq:cco}). 

\noindent 3. Let $\tilde{\Aa},\tilde{\Bb}$ be the closest commuting matrices solving the CCO problem~(\ref{eq:cco}).   
By virtue of their commutativity, $\tilde{\Aa},\tilde{\Bb}$ are jointly diagonalizable, and their joint eigenbasis $\hat{\Uu}$ solves the JADE problem~(\ref{eq:jade}). 
\end{theo}

A big advantage of this approach compared to JADE is the possibility to demand that the closest commuting matrices define valid Laplacians, i.e., restrict the search space to $\mathcal{L}(V,E)$: 
\begin{eqnarray}
C_\mathrm{L}(\Ll_1,\Ll_2) &=& \displaystyle \min_{ \tilde{\Ll}_1, \tilde{\Ll}_2 \in \mathcal{L}(V,E) }  
\sum_{i=1}^2 \| \tilde{\Ll}_i - \Ll_i \|_\mathrm{F}^2  
\,\,\, \text{s.t.} \,\,\, \tilde{\Ll}_1 \tilde{\Ll}_2 = \tilde{\Ll}_2 \tilde{\Ll}_1.  
\label{eq:cco_}
\end{eqnarray} 
%


\section{Heat kernel coupling}

The methods described in Section 3 rely on the assumption of graphs with equal vertex set, which may be too restrictive in many cases. 
More generally, we are given two different graphs $G_k = (V_k,E_k,\bb{W}_k),\, k= 1,2$, where $|V_1|\neq |V_2|$. The correspondence between the vertices is not bijective anymore, but one can consider {\em functional correspondence} $T: \mathcal{F}(V_1) \rightarrow \mathcal{F}(V_2)$, represented by the $|V_2| \times |V_1|$ matrix $\bb{T}$ \cite{ovsjanikov2012functional}. 
%


Let us consider the heat equation~(\ref{eq:heat}) on the graphs $G_1, G_2$. 
We say that the corresponding heat kernels are {\em strongly coupled} if 
the solution of the heat equation on $G_1$ with some initial condition $\bb{f} \neq \mathrm{const}$ and 
the solution of the heat equation on $G_2$ with the corresponding initial condition $\bb{T f}$ coincide under the correspondence:  
\begin{equation}
\bb{T} \bb{H}_1^t \bb{f} = \bb{H}_2^t \bb{T f}, 
\label{eq:coupling}
\end{equation}
for $t \geq 0$. 
The strong coupling condition 
implies that the structure of the graphs is similar, in the sense that heat flows on them in the same way.
\footnote{If the strong coupling condition holds for a set of functions that span the whole $\mathcal{F}(V_1)$, it is equivalent to commutativity of the heat-  and the functional correspondence operators, $\bb{T} \bb{H}_1^t  = \bb{H}_1^t \bb{T}$.  
In the case of bijective correspondence ($|V_1| = |V_2|$ and w.l.o.g. $\bb{T}=\bb{I}$), 
having the strong coupling condition hold for one value of $t>0$ implies that $\bb{L}_1 = \bb{L}_2$ and thus the weighted graphs are isometric \cite{ovsjanikov2010one}.  }
%

If the correspondence $\bb{T}$ is further assumed to be unknown, we have to replace the strong coupling condition~(\ref{eq:coupling}) with a {\em weak coupling}  condition 
\begin{equation}
\bb{f}^\Tr  \bb{H}_1^t \bb{f} = \bb{f}^\Tr \bb{T}^\Tr \bb{H}_2^t \bb{T f}, 
\label{eq:coupling1}
\end{equation}
requiring that the projection of the solution on the corresponding functions is equal. 
Note that while condition~(\ref{eq:coupling}) compares vectors (which requires the knowledge of correspondence $\bb{T}$), condition~(\ref{eq:coupling1}) compares scalars, which does not require the knowledge of the correspondence $\bb{T}$ but rather the pair of corresponding functions $\bb{f}$ and $\bb{T f}$. 
Obviously, condition~(\ref{eq:coupling}) implies~(\ref{eq:coupling1}), but not vice versa.

In this weaker setting, we assume that correspondence $\bb{T}$ is unknown, but we have a set of $q$ {\em corresponding functions} on $V_1$ and $V_2$ represented as columns of matrices $\bb{F} = (\bb{f}_1, \hdots, \bb{f}_q)$ and $\bb{G} = (\bb{g}_1, \hdots, \bb{g}_q)$ such that $\bb{T} \bb{F} = \bb{G}$. 
Writing the weak coupling condition~(\ref{eq:coupling1}) for every pair $\bb{f}_i, \bb{g}_j$, we get the condition on the equality of $q\times q$ matrices 
\begin{equation}
\bb{F}^\Tr  \bb{H}_1^t \bb{F} = \bb{G}^\Tr \bb{H}_2^t \bb{G}, 
\label{eq:coupling2}
\end{equation}
%
which 
we consider on a finite set of the values $t \in \{ t_1, \hdots, t_p\}$. 

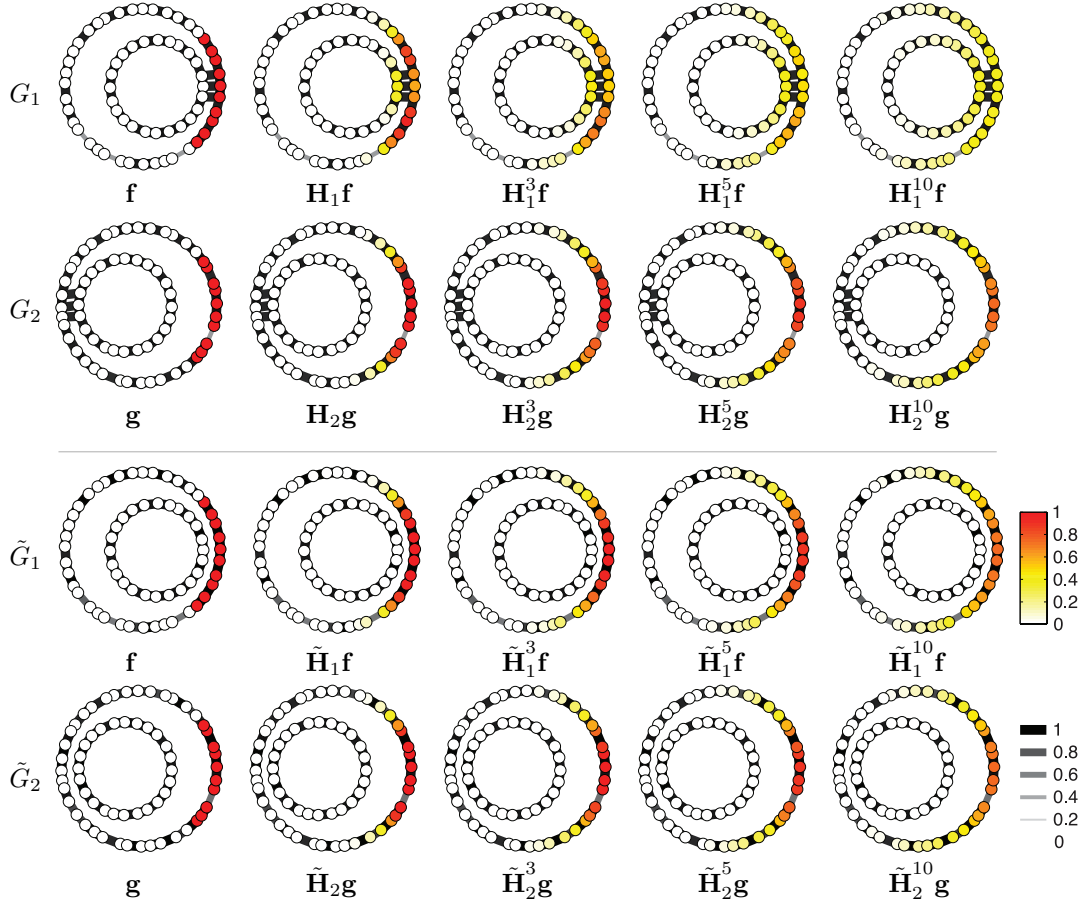
\begin{figure}
\center{
\begin{overpic}
  [width=1\linewidth]{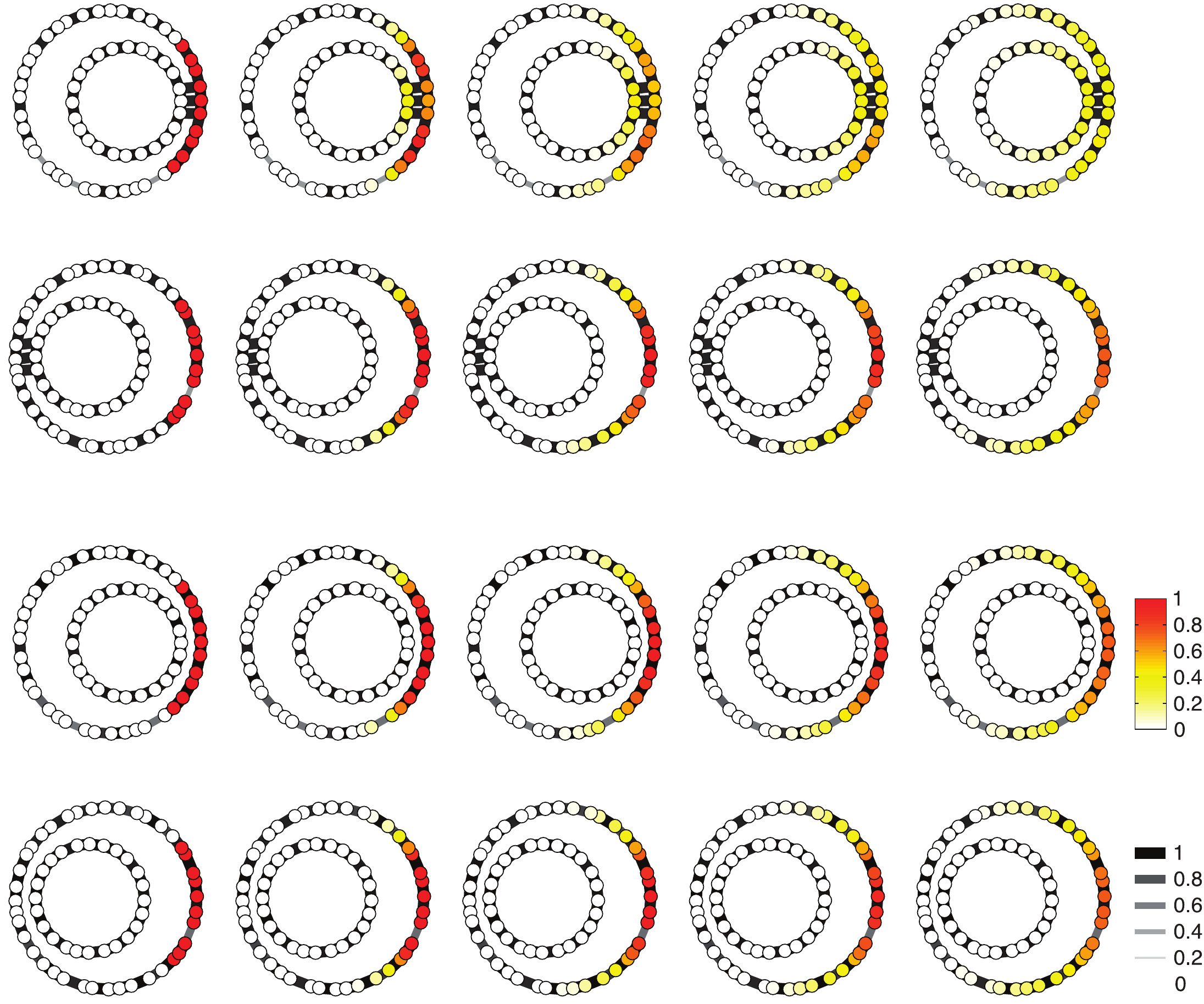}
    \put(-3.75,7){\footnotesize $\tilde{G}_2$}
    \put(-3.75,28.5){\footnotesize $\tilde{G}_1$}
    \put(-3.75,52.5){\footnotesize $G_2$}
    \put(-3.75,73.5){\footnotesize $G_1$}
%
    \put(7.5,63.75){\footnotesize $\bb{f}$}    
    \put(25,63.75){\footnotesize ${\bb{H}}_1 \bb{f}$}    
    \put(44,63.75){\footnotesize ${\bb{H}}_1^{3} \bb{f}$}    
    \put(63,63.75){\footnotesize ${\bb{H}}_1^{5} \bb{f}$}    
    \put(81.5,63.75){\footnotesize ${\bb{H}}_1^{10} \bb{f}$}    
    \put(7.5,42.25){\footnotesize $\bb{g}$}    
    \put(25,42.25){\footnotesize ${\bb{H}}_2 \bb{g}$}    
    \put(44,42.25){\footnotesize ${\bb{H}}_2^{3} \bb{g}$}    
    \put(63,42.25){\footnotesize ${\bb{H}}_2^{5} \bb{g}$}    
    \put(81.5,42.25){\footnotesize ${\bb{H}}_2^{10} \bb{g}$}    
    \put(7.5,18.25){\footnotesize $\bb{f}$}    
    \put(25,18.25){\footnotesize $\tilde{\bb{H}}_1 \bb{f}$}    
    \put(44,18.25){\footnotesize $\tilde{\bb{H}}_1^{3} \bb{f}$}    
    \put(63,18.25){\footnotesize $\tilde{\bb{H}}_1^{5} \bb{f}$}    
    \put(81.5,18.25){\footnotesize $\tilde{\bb{H}}_1^{10} \bb{f}$}    
    \put(7.5,-3.25){\footnotesize $\bb{g}$}    
    \put(25,-3.25){\footnotesize $\tilde{\bb{H}}_2 \bb{g}$}    
    \put(44,-3.25){\footnotesize $\tilde{\bb{H}}_2^{3} \bb{g}$}    
    \put(63,-3.25){\footnotesize $\tilde{\bb{H}}_2^{5} \bb{g}$}    
    \put(81.5,-3.25){\footnotesize $\tilde{\bb{H}}_2^{10} \bb{g}$}    
    %
    \put(1,39.5){\color[rgb]{0.7,0.7,0.7}{\line(1,0){91}} }
\end{overpic}\vspace{3mm}
  \caption{\label{fig:motivation}  \small Solution of the heat equation on the original graphs $G_1, G_2$ (first and second rows) using corresponding initial conditions $\bb{f}$ and $\bb{g} = \bb{T f}$ (left) does not satisfy the coupling condition, $\bb{H}_1^t \bb{f} \neq \bb{T} \bb{H}_2^t \bb{g}$. 
  On graphs $\tilde{G}_1, \tilde{G}_2$ modified using our HKC procedure (third and fourth rows), the heat kernels are approximately coupled, $\tilde{\bb{H}}_1^t \bb{f} \approx \bb{T} \tilde{\bb{H}}_2^t \bb{g}$.
  In this and in following figures, the values of a function defined on the graph vertices are shown in color; the graph edge weights are shown in line width and gray shade. 
}
}
\end{figure}

Typically, two different graphs will have their heat kernels uncoupled, violating the coupling conditions (see example in Figure~\ref{fig:motivation} (top), where different behavior of the heat equation stems from topological noise).  
The problem of {\em heat kernel coupling} (HKC) treated in this paper is how to {\em minimally} modify the Laplacians of the graphs to make the respective heat kernels (approximately) satisfy the weak coupling condition by enforcing~(\ref{eq:coupling2});  
%
%
in Figure~\ref{fig:motivation} (bottom) such a modification amounts to disconnecting the rings in both graphs.

%
%
%
%
%

Our HKC problem bears resemblance to the CCO problem described in Section 3: 
we are looking for new graphs $\tilde{G}_k = (V_k,E_k, \tilde{\bb{W}}_k)$ 
%
with respective adjacency matrices $\tilde{\Ww}_k$, such that the new Laplacians  $\tilde{\Ll}_k$ are as close as possible to the original $\Ll_k$, and the 
corresponding new heat operators $\tilde{\bb{H}}_k^t = e^{-t \tilde{\bb{L}}_k}$ are as coupled as possible,  
%
%
\begin{eqnarray}
\label{eq:hkc}
\displaystyle \min_{ \tilde{\bb{L}}_k \in \mathcal{L}(V_k, E_k) }  
\sum_{k=1}^2 \| \tilde{\Ll}_k - \Ll_k \|_\mathrm{F}^2 
+ 
\alpha \sum_{m=1}^p \| \bb{F}^\Tr e^{-t_m \tilde{\bb{L}}_1} \bb{F} - \bb{G}^\Tr e^{-t_m \tilde{\bb{L}}_2} \bb{G} \|_\mathrm{F}^2. 
\end{eqnarray}

It is important to observe the following limit case: 
for graphs $G_k = (V,E,\Ww_k)$ with equal vertex and edge sets discussed in Section 3, we have bijective correspondence between the entries of the heat operators $\tilde{\bb{H}}_1, \tilde{\bb{H}}_2$, implying  $\Ff = \bb{G} = \bb{I}$. 
In the limit $\alpha \rightarrow \infty$, we have $\tilde{\bb{H}}^t_1 = \tilde{\bb{H}}^t_2$, from which it follows that 
$\tilde{\bb{L}}_1 = \tilde{\bb{L}}_2 = \tilde{\Ll}$. Thus, the HKC problem~(\ref{eq:hkc}) boils down to the simple Laplacian averaging~(\ref{eq:avg}), and can be considered an extension of this technique to the setting where one cannot straightforwardly average Laplacians since the correspondence between the graphs is not given. 

\section{Numerical optimization}

We parametrize our problem through the adjacency matrix $\tilde{\bb{W}}_k(\bb{u}_k)$, where the edge weights $\bb{u}_k = (u^k_1,\hdots, u^k_{|E_k|})$ are defined according to (\ref{eq:wparam}). 
Problem~(\ref{eq:hkc}) can be rewritten as 
\begin{eqnarray}
\label{eq:hkc1}
\displaystyle \min_{ \bb{u}_1, \bb{u}_2 \geq 0 }  
\sum_{k=1}^2 \| \tilde{\Ll}_k(\bb{u}_k) - \Ll_k \|_\mathrm{F}^2 
+ 
\alpha \sum_{m=1}^p \| \bb{F}^\Tr e^{-t_m \tilde{\bb{L}}_1(\bb{u}_1)} \bb{F} - \bb{G}^\Tr e^{-t_m \tilde{\bb{L}}_2(\bb{u}_2)} \bb{G} \|_\mathrm{F}^2. 
\end{eqnarray}
Its solution is carried out using standard optimization techniques, requiring the gradient of the cost function. 

We differentiate the cost function~(\ref{eq:hkc1}) w.r.t. the edge weights $\tilde{w}^k_{ij} : (i,j) \in E_k$ constituting the vectors $\bb{u}_k$, accounting for the symmetric structure of $\tilde{\bb{W}}_k$. 
%
%
The gradient of the distance term is given by 
\[
\frac{\partial \| \tilde{\Ll}_k - \Ll_k \|_\mathrm{F}^2} {\partial \tilde w^k_{ij}}	= 2(\bb{O}_k + \bb{O}_k^\Tr-2(\tilde{\Ll}_k - \Ll_k)   )_{ij}
\]
where 
$\bb{O}_k = (\mathrm{diag}(\tilde{\Ll}_k - \Ll_k), \hdots, \mathrm{diag}(\tilde{\Ll}_k - \Ll_k) )$ is an $|V_k|\times |V_k|$ matrix with equal columns containing the diagonal of $\tilde{\Ll}_k - \Ll_k$. 

The gradient of the coupling term is computed by applying the chain rule several times, as follows. 
First, let 
\begin{eqnarray*}
\tfrac{\partial }{\partial w^k_{ij}}\bb{\tilde{L}}_k &=&
\left(
	\begin{array}{ccccc}
	.&.&.&.&.\\
	.&+1&.&-1&.\\
	.&.&.&.&.\\
	.&-1&.&+1&.\\
	.&.&.&.&.\\
	\end{array}\right)
\end{eqnarray*}
be a $|V_k| \times |V_k|$ matrix containing only four non-zero elements in its $i$th and $j$th row and column. 
%
Second, for each $k=1,2$ and $m=1,\hdots, p$ compute the $2|V_k| \times 2|V_k|$ matrix exponent 
\begin{eqnarray*}
\bb{H}^{ij}_{k,m}&=&\exp\left(
\begin{array}{cc}
-t_m\tilde {\Ll}_k&-t_m \tfrac{\partial }{\partial w^k_{ij}}\bb{\tilde{L}}_k \\
0&-t_m\tilde{\Ll}_k\\
\end{array}\right)
\end{eqnarray*}
and extract its $|V_k| \times |V_k|$ upper right block, which we denote by $\hat{\bb{H}}^{ij}_{k,m}$. 
Finally, 
\begin{eqnarray*}
\frac{\partial }{\tilde{w}^1_{ij} }\| \Ff^\Tr e^{-t_m \tilde{\Ll}_1} \Ff - \bb{G}^\Tr e^{-t_m \tilde{\Ll}_2} \bb{G} \|_\mathrm{F}^2 &=& 2\tr( \Ff(\Ff^\Tr e^{-t_m \tilde{\Ll}_1} \Ff-\bb{G}^\Tr e^{-t_m \tilde{\Ll}_2} \bb{G})\Ff^\Tr \hat{\bb{H}}_{1,m}^{ij}); \\
\frac{\partial }{\tilde{w}^2_{ij} }\| \Ff^\Tr e^{-t_m \tilde{\Ll}_1} \Ff - \bb{G}^\Tr e^{-t_m \tilde{\Ll}_2} \bb{G} \|_\mathrm{F}^2 &=& 
2\tr( \bb{G}(\bb{G}^\Tr e^{-t_m \tilde{\Ll}_2} \bb{G} - \Ff^\Tr e^{-t_m \tilde{\Ll}_1} \Ff)\bb{G}^\Tr \hat{\bb{H}}_{2,m}^{ij}). 
\end{eqnarray*}

%
%
%

\section{Results}

In this section, we demonstrate our HKC approach on several synthetic and real datasets coming from shape analysis, manifold learning, and pattern recognition problems. 
The experiments closely follow our previous work \cite{cco}, and their leitmotif is, given two datasets representing similar objects in somewhat different ways, to reconcile the information of the two modalities producing a single consistent representation. 
We should stress that though we know the groundtruth correspondence between the vertices of the graphs representing different modalities, we are not using it in our HKC problem. Instead, we only assume to be given few corresponding functions $\bb{F}, \bb{G}$ that are used to couple the heat kernels.

In all the experiments, we used unnormalized Laplacians (\ref{eq:laplacian}) constructed with Gaussian weights. 
We used $\alpha=10^6$ in the cost~(\ref{eq:hkc1}). 
Optimization was performed using MATLAB optimization toolbox.

{\bf Circles. } We used two graphs shaped as two eccentric circles , containing 64 points and having different connectivity (Figure~\ref{fig:motivation}, top). We used four corresponding functions in the HKC optimization. 
The closest Laplacians that produce coupled heat kernels result in edge weights shown in Figure~\ref{fig:motivation} (bottom): the optimization performs a `surgery' disconnecting the inconsistent connections and producing two connected components. 

\begin{figure}
\center{
\begin{overpic}
  [width=1\linewidth]{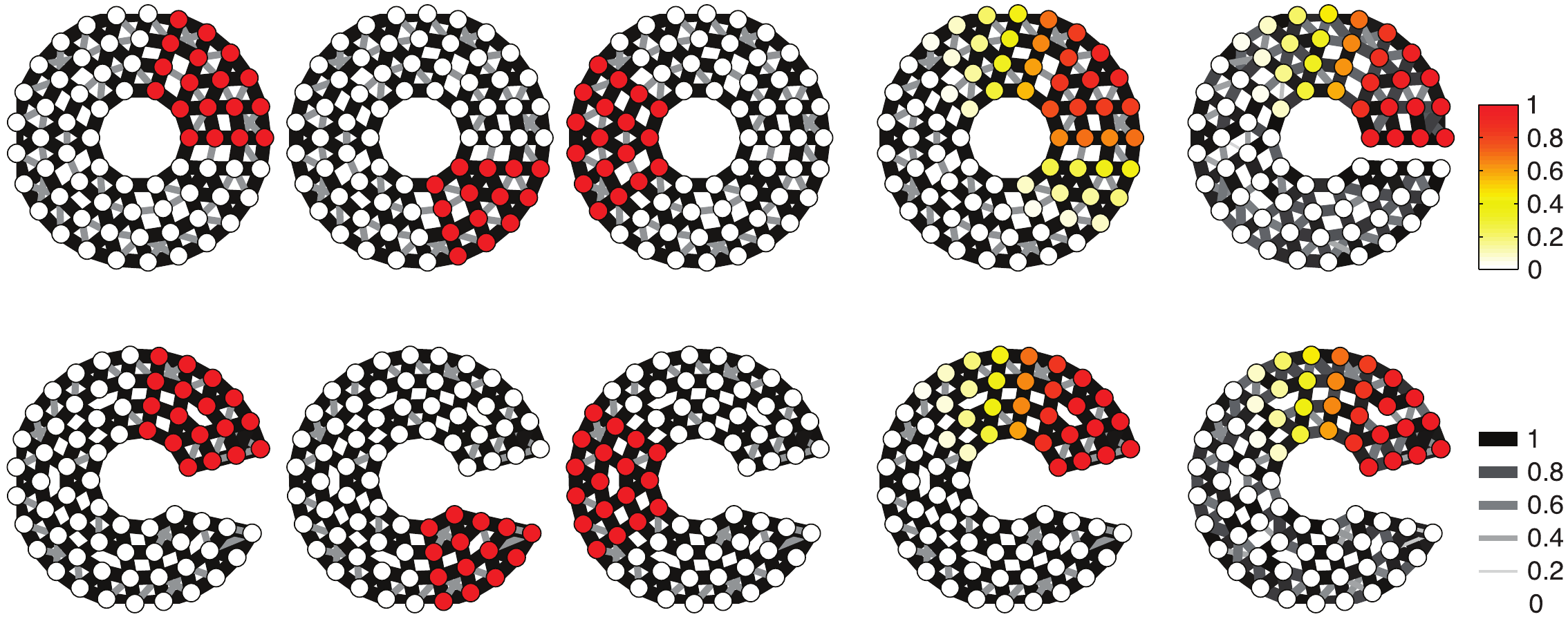}  
      \put(8,19.25){\footnotesize $\bb{f}_{1}$}
      \put(26,19.25){\footnotesize $\bb{f}_{2}$}
      \put(44,19.25){\footnotesize $\bb{f}_{3}$}     
      \put(62,18.8){\footnotesize $\bb{H}^{10}_1\bb{f}_{1}$}       
      \put(82,18.8){\footnotesize $\tilde{\bb{H}}^{10}_1\bb{f}_{1}$}       
      \put(8,-2.5){\footnotesize $\bb{g}_{1}$}
      \put(26,-2.5){\footnotesize $\bb{g}_{2}$}
      \put(44,-2.5){\footnotesize $\bb{g}_{3}$}     
      \put(62,-3.1){\footnotesize $\bb{H}^{10}_2\bb{g}_{1}$}       
      \put(82,-3.1){\footnotesize $\tilde{\bb{H}}^{10}_2\bb{g}_{1}$}
\end{overpic}\vspace{2mm}
  \caption{\label{fig:ring}  \small 
  {\em Ring} experiment. 
  Left: coupling functions;  
 Right:  Solution of the heat equation at $t=10$ on the original graphs $G_1, G_2$ (fourth column) and the modified graphs $\tilde{G}_1, \tilde{G}_2$ (fifth column) using $\bb{f}_1, \bb{g}_1$ as initial conditions.   
  The modification amounts to cutting the ring $G_1$ (top right).
}
}
\end{figure}

{\bf Ring. } 
We used a ring and a cracked ring sampled at 70 points and connected using four nearest neighbors (Figure~\ref{fig:ring}, top and bottom). Three functions only were used for coupling (Figure~\ref{fig:ring}, three leftmost columns).  
Because of the topological difference, the behavior of the heat flow differs dramatically (Figure~\ref{fig:ring}, fourth column from left)
The HKC optimization cuts the connections in the first graph, making the two rings topologically equivalent and resulting in the same heat flow 
(Figure~\ref{fig:ring}, rightmost)

\begin{figure}
\center{
\begin{overpic}
  [width=1\linewidth]{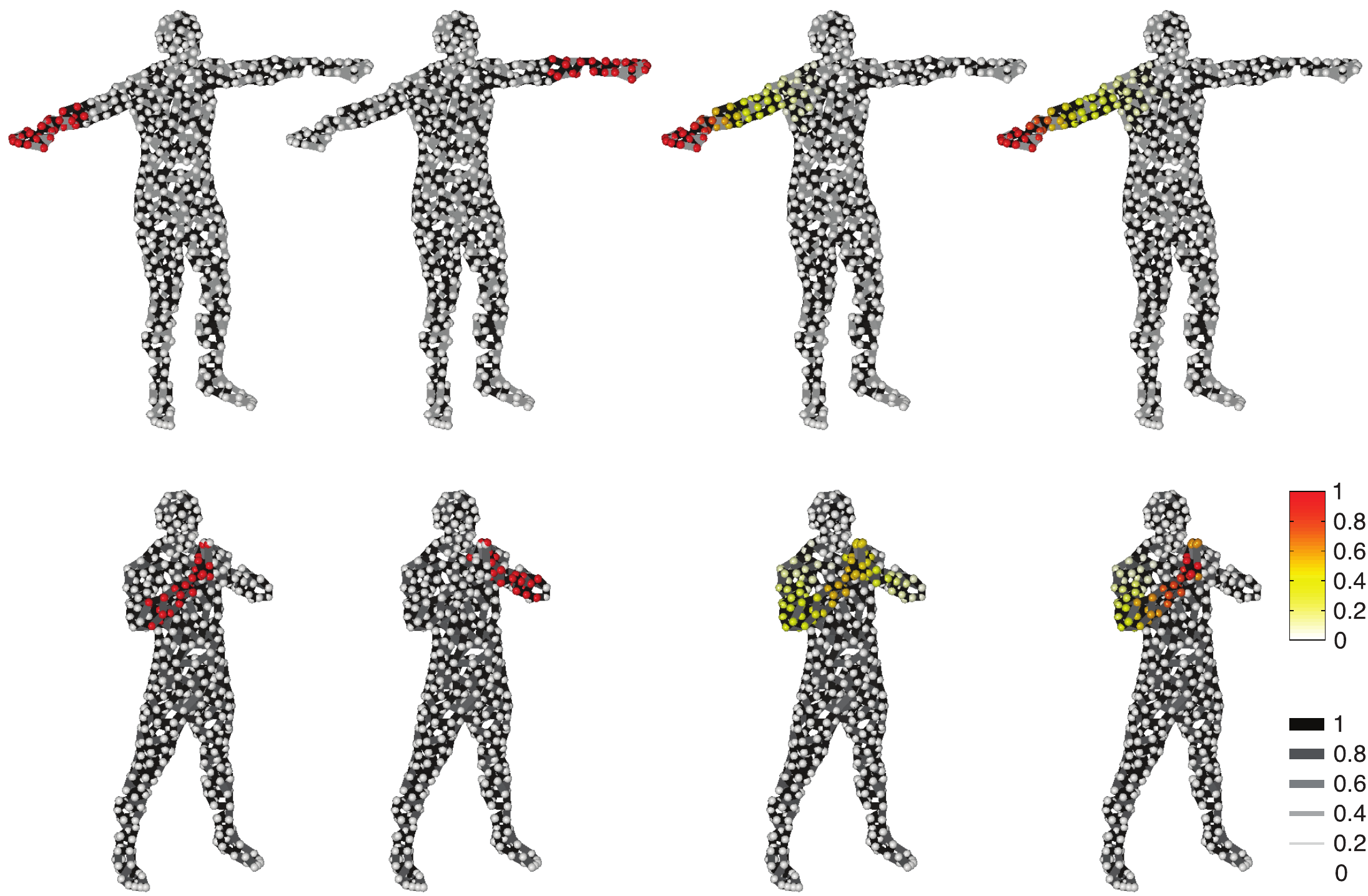}  
      \put(14,32){\footnotesize $\bb{f}_{1}$}
      \put(35,32){\footnotesize $\bb{f}_{2}$}
      \put(14,-1.5){\footnotesize $\bb{g}_{1}$}
      \put(35,-1.5){\footnotesize $\bb{g}_{2}$}
      \put(61.25,32){\footnotesize $\bb{H}^{20}_1\bb{f}_{1}$}
      \put(85.25,32){\footnotesize $\tilde{\bb{H}}^{20}_1\bb{f}_{1}$}
      \put(60.5,-1.5){\footnotesize $\bb{H}^{20}_2\bb{g}_{1}$}
      \put(85.0,-1.5){\footnotesize $\tilde{\bb{H}}^{20}_2\bb{g}_{1}$}
\end{overpic}\vspace{2mm}
  \caption{\label{fig:man}  \small 
  {\em Man} experiment. 
  Left: coupling functions; 
 Right:  Solution of the heat equation at $t=20$ on the original graphs $G_1, G_2$ (third column) and the modified graphs $\tilde{G}_1, \tilde{G}_2$ (fourth column) using $\bb{f}_1, \bb{g}_1$ as the initial conditions.  
}
}
\end{figure}

{\bf Man. } 
We used two poses of the human shape from the TOSCA dataset \cite{bronstein2008ngn}, uniformly sampled at 500 points and connected using five nearest neighbors. The resulting graphs have different topology (the hands are connected or disconnected, compare Figure~\ref{fig:man} top and bottom), resulting in a very different heat flow. 
Two functions were used for coupling (Figure~\ref{fig:man}, two leftmost columns) in our HKC problem; our optimization disconnects these links (Figure~\ref{fig:man}, right) making the heat flow in both cases behave similarly.

{\bf NUS. } We used a subset of the NUS-WIDE dataset \cite{nus-wide-civr09} containing images (represented by 64-dimensional color histograms) and their text annotations (represented by 1000-dimensional distributions of most frequent tags) from seven classes. 
The classes were selected on purpose in order to be ambiguous in different modalities: for example, in the Tags modality {\em underwater tigers} can be similar both to {\em tigers} and {\em water animals}, as they share many tags. 
On the other hand, in the Color modality {\em tigers} may be similar to the class of {\em nature}, containing images with orange-yellow autumn colors  \cite{eynard2012multimodal}. 

In each modality, we used Laplacians with Gaussian weights and 25 nearest neighbors, computed with self-tuning scales. 
Seven functions were used for coupling. 
We computed diffusion distances~(\ref{eq:diffdist}) on the original and the modified graphs, and used them to rank the dataset entries in a leave-one-out retrieval experiment.  
Retrieval performance was evaluated using {\em mean average precision} $mAP = \sum_{r=1}^R P(r) \cdot rel(r)$, where $rel(r)$ is the relevance of a given rank (one if it belongs to the same class of the query and zero otherwise), $R$ is the number of retrieved results, and $P(r)$ is {\em precision at $r$}, defined as the percentage of relevant results in the first $r$ top-ranked retrieved matches. Respectively, {\em recall} $R(r)$ is defined as the percentage of relevant results in the first $r$ top-ranked retrieved matches out of all items belonging to the query class.

\begin{figure}
\center{
\begin{overpic}
  [width=0.75\linewidth]{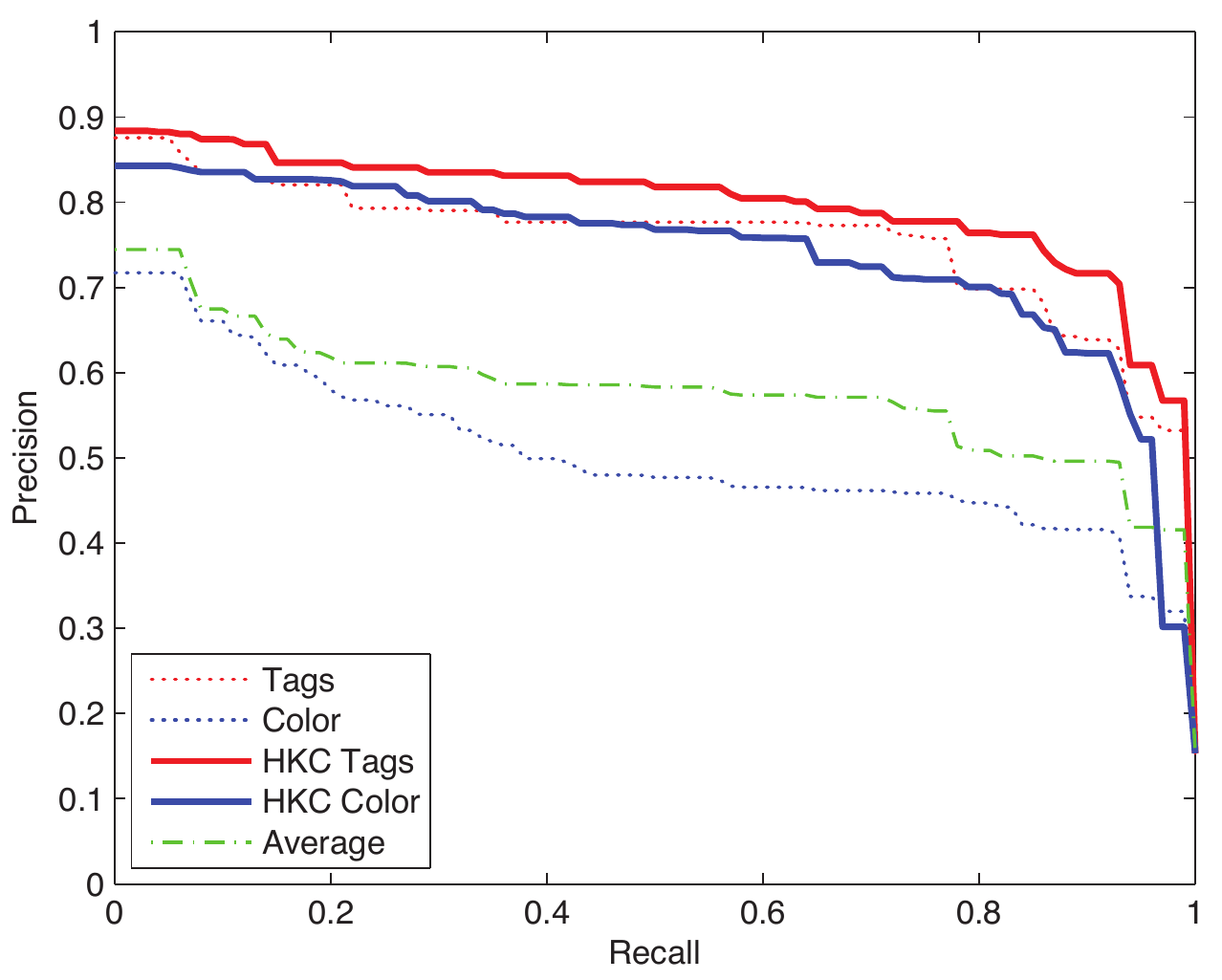}  
\end{overpic}\vspace{0mm}
  \caption{\label{fig:nus-roc}  \small 
Precision-recall curve of the  {\em NUS} experiment, using diffusion distances $d_1$ and $d_2$ computed from uni-modal Laplacians (Tags modality (dotted red) and Colors modality, (dotted blue), respectively), diffusion distances $\tilde{d}_1$ and $\tilde{d}_2$ computed from the HKC-modified Laplacians (solid red and blue, respectively). 
Result of Laplacian averaging (green dash-dot) requires bijective correspondence between the data in the two modalities and is shown for reference only. 
In all cases, time scale $t=1.25$ is used. 
}
}
\end{figure}

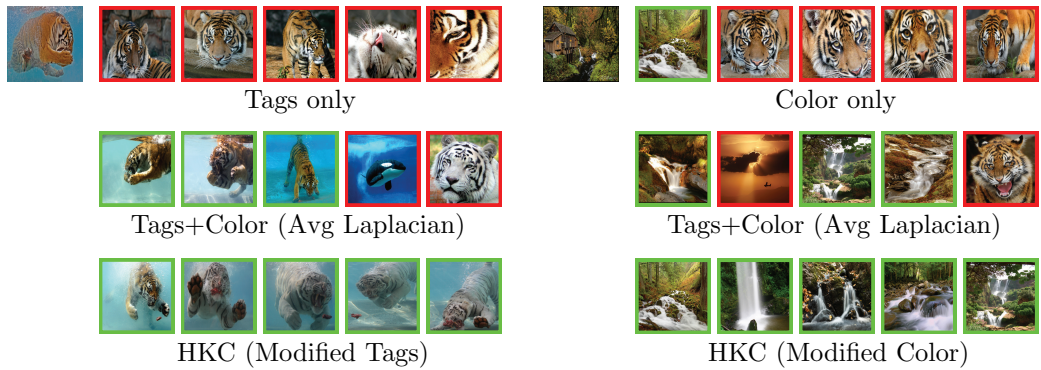
\begin{figure}
\center{
\begin{overpic}
  [width=1\linewidth]{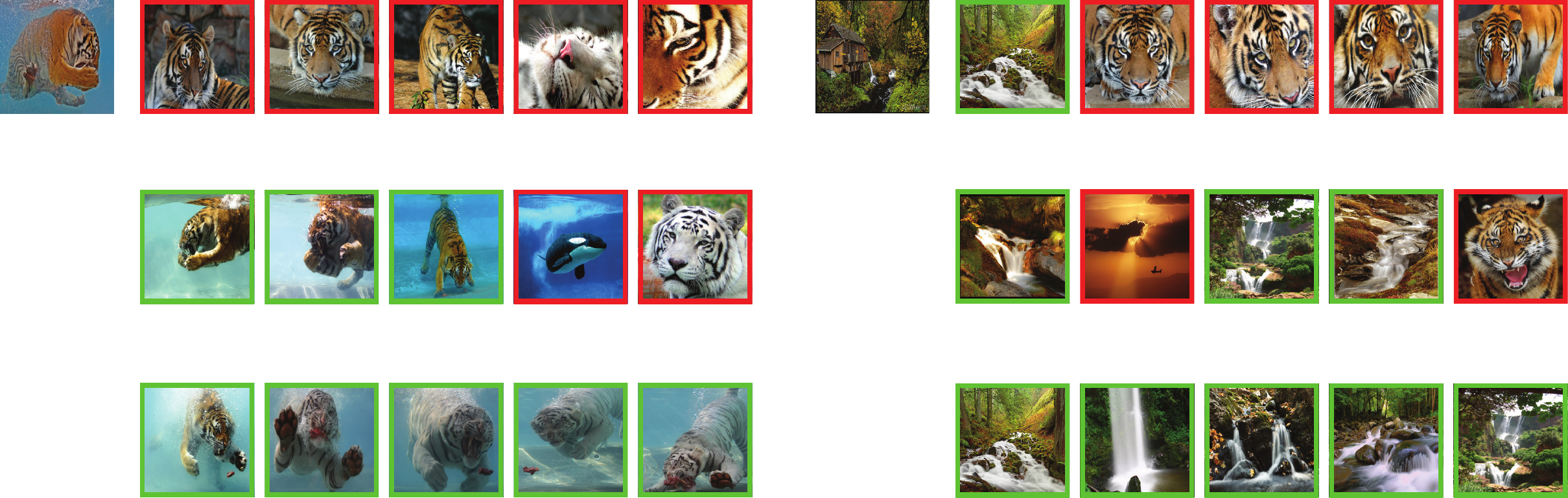}  
      \put(23,21.9){\footnotesize Tags only}
      \put(16.5,-2.6){\footnotesize HKC (Modified Tags)}
      \put(12,9.8){\footnotesize Tags+Color (Avg Laplacian)}
      \put(74.5,21.9){\footnotesize Color only}
      \put(68,-2.6){\footnotesize HKC (Modified Color)}
      \put(64,9.8){\footnotesize Tags+Color (Avg Laplacian)}
\end{overpic}\vspace{2mm}
  \caption{\label{fig:nus-nn}  \small 
  {\em NUS} experiment. 
Five nearest neighbors to two queries measured using diffusion distance with $t=1.25$, using single modality (first row; Tags: left, Color: right), multiple modality by Laplacian averaging (second row) and HK coupling (third row). 
Matches belonging to the same class of the query are marked with green. 
Using Tags modality only, the {\em swimming tiger} query (tagged as {\em underwater, tiger, animal, zoo}) is confused with {\em tigers} (tagged as {\em animal, zoo, cat, tiger, nature}).   
Using Color modality only, the orange-yellow colored {\em nature} query is confused with similarly colored {\em tigers}. 
}
}
\end{figure}

Figure~\ref{fig:nus-roc} shows the precision-recall curve of different methods, and Table~\ref{tab:nus} summarizes the mean average precision. We can see that after HKC optimization, performance increases significantly, outperforming each modality on its own. 
Figure~\ref{fig:nus-nn} shows examples of first matches corresponding to ambiguous queries. 
For reference only, we show the performance of Laplacian averaging, which however relies on bijective correspondence between the graphs (which is not used in our HKC problem) and is thus not directly comparable.

\begin{table*}[htdp]\small
\begin{center}
\begin{tabular}{l c |cc}
   Method & $t$ 
   & \small Precision@5  & \small mAP  \\
\hline
%
 \multirow{3}{*}{\hspace{-2mm}\small Tags only \hspace{-2mm} } 
  & 0.75     & \small 82.3 \% & \small 78.4 \%  \\
 & 1.0         & \small 81.2 \% & \small 77.0 \%  \\
& 1.25     & \small 79.7 \% & \small 76.2 \% \\
\cline{1-4}
 \multirow{3}{*}{\hspace{-2mm}\small Color only \hspace{-2mm} } 
  & 0.75     & \small 61.8 \% & \small 55.7 \% \\
 & 1.0         & \small 61.6 \% & \small 53.5 \% \\
& 1.25     & \small 59.6 \% & \small 51.5 \% \\
\cline{1-4}
 \multirow{3}{*}{\hspace{-2mm}\small HKC Tags \hspace{-2mm} } 
  & 0.75     & \small 87.3 \% & \small 82.2 \% \\
 & 1.0         & \small 86.3 \% & \small 81.5 \% \\
& 1.25     & \small 84.4 \% & \small 80.3 \% \\
\cline{1-4}
 \multirow{3}{*}{\hspace{-2mm}\small HKC Color \hspace{-2mm} } 
  & 0.75     & \small 83.2 \% & \small 76.2 \% \\
 & 1.0         & \small 82.3 \% & \small 75.6 \% \\
& 1.25     & \small 80.6 \% & \small 74.6 \% \\
\cline{1-4}
 \multirow{3}{*}{\hspace{-2mm}\small Average \hspace{-2mm} } 
 & 0.75     & \small 68.7 \% & \small 64.2 \% \\
 & 1.0         & \small 66.5 \% & \small 61.0 \% \\
& 1.25     & \small 63.6 \% & \small 58.8 \% \\
\hline
\end{tabular}\vspace{-2mm}
\end{center}
\caption{\label{tab:nus} Retrieval performance on the NUS dataset using diffusion distances with different time scale $t$. 
 }
\end{table*}

\section{Conclusions}

We showed the heat kernel coupling problem, whereby we seek to minimally modify a pair of Laplacians to make the corresponding heat kernels to become coupled, such that the solution of a heat equation on two graphs behaves consistently. 
This problem generalizes simple Laplacian averaging to the setting when the correspondence between the two graphs is unknown.

\section{Acknowledgement}

This research was supported by the ERC Starting Grant No. 307047 (COMET).



\bibliographystyle{plain}\small
\bibliography{laplacians.bib}

\end{document}